\documentclass{article}
\usepackage{ijcai22}

\usepackage{times}
\usepackage{soul}
\usepackage{url}
\usepackage[hidelinks]{hyperref}
\usepackage[utf8]{inputenc}
\usepackage[small]{caption}
\usepackage{graphicx}
\usepackage{amsmath}
\usepackage{amsthm}
\usepackage{booktabs}
\usepackage{algorithm}
\usepackage{algorithmic}
\usepackage{amsfonts}       
\usepackage{nicefrac}       
\usepackage{float}
\usepackage{gensymb}
\usepackage{subcaption}
\usepackage{xcolor}

\usepackage{arydshln}
\title{Decentralized adaptive clustering of deep nets is beneficial for client collaboration}

\author{
Edvin Listo Zec$^{1,2}$
\and
Ebba Ekblom$^1$\and
Martin Willbo$^1$\and\\
Olof Mogren$^1$\And
Sarunas Girdzijauskas$^{1,2}$
\affiliations
$^1$RISE Research Institutes of Sweden\\
$^2$KTH Royal Institute of Technology
\emails
edvin.listo.zec@ri.se
}

\begin{document}

\maketitle

\begin{figure*}[t]
    \centering
    \begin{subfigure}[t]{0.3\linewidth}
    \includegraphics[width=\columnwidth]{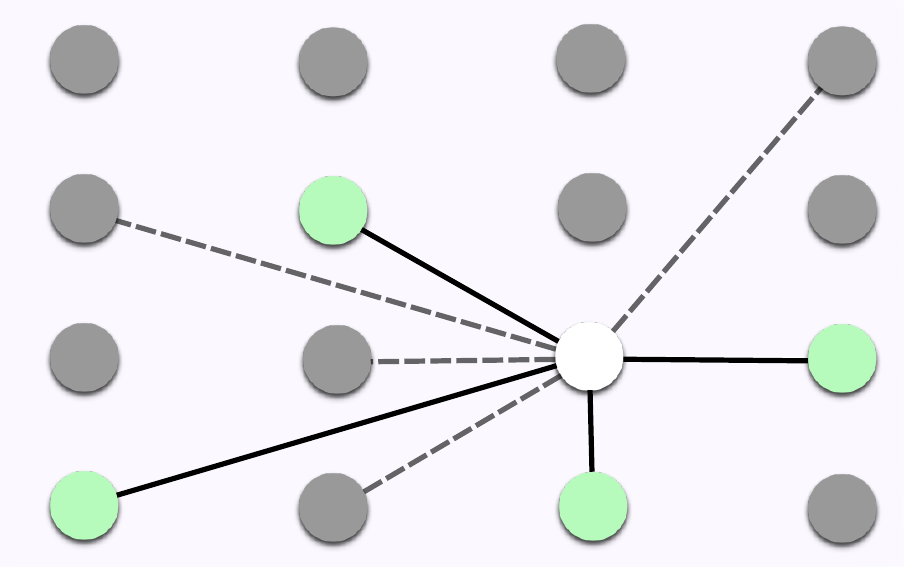}
    \subcaption{\small{Communication round zero. The client has a uniform probability distribution over all other clients in the network. At this point in time the communication strategy is equivalent to random communication.}}
    \label{fig:alg_start}
    \end{subfigure} ~
    \begin{subfigure}[t]{0.3\linewidth}
    \includegraphics[width=\columnwidth]{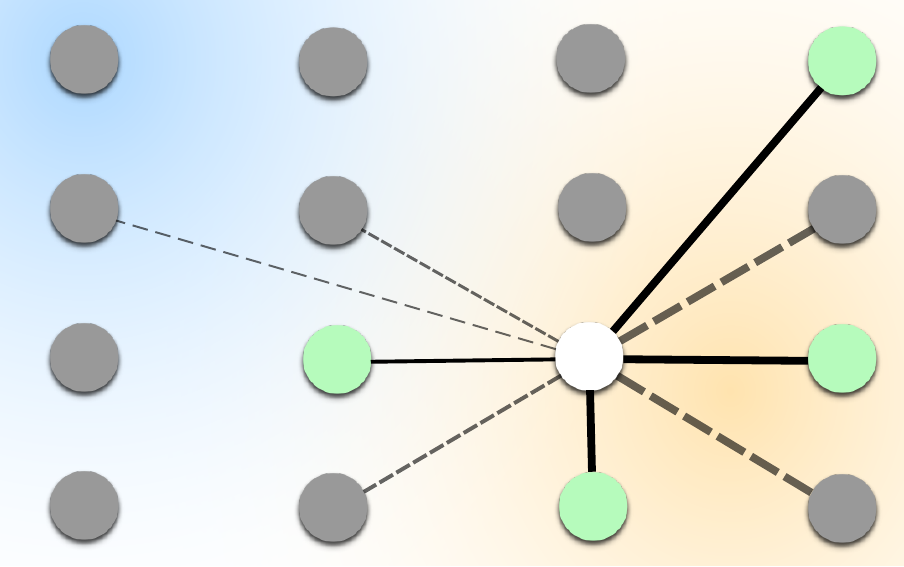}
    \subcaption{\small{After a couple of communication rounds the client's probability distribution over all other clients has been updated based on the similarity measure. A client topology is taking form.}}
    \label{fig:alg_mid}
    \end{subfigure} ~
    \begin{subfigure}[t]{0.3\linewidth}
    \includegraphics[width=\columnwidth]{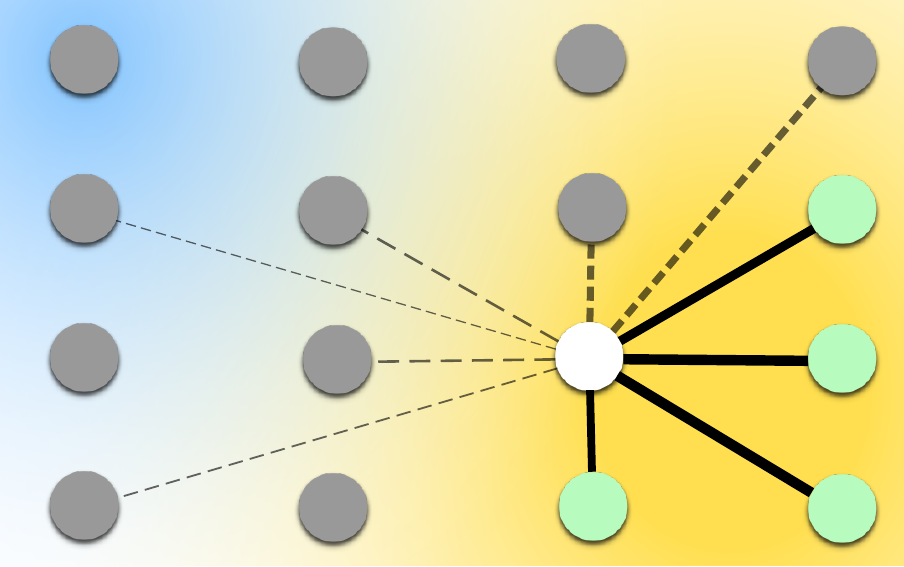}
    \subcaption{\small{The client's probability distribution over all other clients has converged. The communication strategy is now similar to an oracle sampling only within the cluster identity of the client. 
    }}
    \label{fig:alg_end}
    \end{subfigure}
    \caption{\small{Visualization of three consecutive points in time (communication rounds) for DAC. The client in focus (white) selects other clients (green) to communicate with, indicated by solid edges. The probability to do so is indicated by edge thickness. Note that there is a non-zero probability assigned to all other clients in the network, indicated by dashed lines, again with thickness proportional to the probability. For visual clarity only some of these have been explicitly drawn out. The background color indicates the convergence to a network topology from the perspective of client similarity, see equation \eqref{eq:similarity}.}}

    \label{fig:visualization}
\end{figure*}

\begin{abstract}

We study the problem of training personalized deep learning models in a decentralized peer-to-peer setting, focusing on the setting where data distributions differ between the clients and where different clients have different local learning tasks. We study both covariate and label shift, and our contribution is an algorithm which for each client finds beneficial collaborations based on a similarity estimate for the local task. Our method does not rely on hyperparameters which are hard to estimate, such as the number of client clusters, but rather continuously adapts to the network topology using soft cluster assignment based on a novel adaptive gossip algorithm. We test the proposed method in various settings where data is not independent and identically distributed among the clients. The experimental evaluation shows that the proposed method performs better than previous state-of-the-art algorithms for this problem setting, and handles situations well where previous methods fail.


\end{abstract}

\section{Introduction}


The study of machine learning on decentralized data has been of interest for a long time and has received much attention in the last decades. With the spread of smartphones and other mobile devices, an immense increase of data collected has highlighted the need for responsible mining. Modern mobile devices are equipped with sensors that can collect a multitude of data streams related to people's everyday lives, providing a promising source of training data for smart services. However, such data requires special care not only due to increasing privacy concerns among users but also due to regulatory privacy laws. Decentralized machine learning is relevant in many related research communities, such as data mining and telecommunication.
The overarching goal is to efficiently learn from and analyze data distributed among several clients, without the data leaving the clients. 
The aforementioned developments have given rise to a need for algorithms that can efficiently leverage large volumes of data residing on the collecting devices without compromising privacy.

Distributed data collection unlocks the potential of data that would otherwise be difficult or impossible to obtain, but may also lead to heterogeneity between the different datasets.
Different clients often have different types of biases and their data may not always be independently and identically distributed (iid). 
%
This results in differing client data distributions (the non-iid data paradigm), which hinders efficient training of deep learning models \cite{kairouz2021advances}.

In \textit{decentralized learning}, no central server exists and the clients communicate through a peer-to-peer network. The central server in a centralized setup such as federated learning \cite{mcmahan2017communication} is a potential point of weakness: it could fail or be maliciously attacked, leading to failure for the distributed learning. A decentralized system is inherently robust to this. In such a framework, the clients follow a communication protocol in order to leverage each other's data privately. A commonly used protocol is the gossip framework \cite{kempe2003gossip}, where clients communicate and aggregate their models randomly. Gossip based learning between clients works well for the emergence of a single (global) model at each participating node. However, in a non-iid setting, such convergence to a one-size-fits-all solution might be sub-optimal, and even detrimental for some clients.

How to train a deep learning model in a non-iid data setting is an active area of research in the federated learning literature. However, this is a relatively understudied problem in the decentralized setting \cite{kairouz2021advances}. In this work, we propose a solution to the decentralized distributed learning problem where each client adaptively detects which peers to collaborate with during every communication round. The solution is based on a similarity score defined as the inverse of the training loss of one client's model on another client's data. This promotes initiating collaborations which are the most beneficial to the learning of each client. We name this solution \textit{Decentralized adaptive clustering (DAC)} and demonstrate with a thorough experimental investigation that DAC effectively identifies clusters in the network topology and thus achieves strong results in all settings examined. In some experiments, we even outperform the oracle solution which is given complete and perfect information about which peers have the same data distribution.

\section{Related work}

Learning personalized models when data is heterogeneously distributed over the clients is a problem setting studied in both federated learning and decentralized learning.

Gossip learning is a peer-to-peer communication protocol which has been studied in many different machine learning settings \cite{kempe2003gossip,boydgossip,ormgossip}.
Most work on Gossip learning has focused on convex optimization rather than the non-convex optimization associated with training deep neural networks. The first decentralized work on gossip-based optimization for non-convex deep learning studied convolutional neural networks (CNNs) \cite{blotgossip}. The authors showed experimentally that high accuracies with low communication costs are achievable with a decentralized and asynchronous framework. Gossip learning however is not suitable for non-iid settings where several distinct learning objectives may be present as the protocol does not take into consideration which peers may be beneficial for communication during training.

In previous work on federated learning, \cite{ghosh2020efficient} showed that having several global models in a federated setup improves performance over a single global model when client data is not identically distributed. The goal was to have clients with similar learning objectives assigned to their own cluster, then aggregate parameters internally and create one global model for each cluster. The algorithm thus alternates between client cluster assignment and loss minimization. Cluster identity is estimated by empirical loss given current global models evaluated on client data. However, a drawback of this method is that it relies on hyperparameters such as the number of clusters (i.e. the number of global models to learn). Additionally, the experiments are limited to hard cluster assignment, and cannot handle soft clustering. Further, the client data heterogeneity is limited to covariate-shift.

In previous work on decentralized learning, \cite{onoszko2021decentralized} developed an algorithm based on training loss which identifies similar clients. While their experimental results are strong, they also limit themselves to uniform cluster sizes and only study covariate-shift. Further, their method is based on two steps: 1) a cluster identification step, and 2) random communication within the identified clusters. 
The approach requires a hyperparameter search for the length of the search for similar clients, and the hard cluster assignment has no way of handling varying degrees of similarity between local training datasets.
It is also not obvious that the optimal strategy is to only communicate within your own cluster - especially if you belong to a cluster with few members. In this work, we also explore the robustness of different algorithms in such cases.

The approach proposed in this paper is a decentralized learning algorithm that continuously adapts the client communication topology every communication round based on a client similarity metric. Compared to previous work on clustered federated learning this approach does not rely on specifying the number of clusters a priori. Similarly, our method eliminates the need for a cluster identification step as used in \cite{onoszko2021decentralized}. 
We show that a soft cluster assignment based on a client similarity measure
instead of just finding and communicating with clients with the same learning objective makes it possible to beat an all-knowing oracle which only communicates within pre-defined clusters. This is especially the case in situations where clusters consist of only a few clients.  
\vspace{-5pt}
\section{Method}
We assume a decentralized learning setting with $K$ clients which are able to communicate in a communication network for $T$ number of rounds. Each client $i$ has a \textit{private} training data distribution $\mathcal{D}_i(x,y)$ over input features $x$ and labels $y$. Each client also has a deep neural network $f_i$ with model parameters $w_{i}\in\mathbb{R}^{d}$. Let $\ell(f_i(w_{i};x),y) : \mathbb{R}^{d} \to \mathbb{R}$ be the loss as a function of the model parameters $w_{i}$ and data $(x,y)$ for client $i$. The principal problem of this work is then to define a decentralized learning algorithm which leverages the client's own data and models of other clients over the $T$ communication rounds in such a way that the resulting $w_{i}$ are a solution to the optimization problem described in \eqref{eq:opt} below, which corresponds to each client $i$ learning its local task:

\begin{equation}
\label{eq:opt}
\min_{w_{i}\in\mathbb{R}^{d}} \mathcal{L}(f_i(w_{i})) := \min_{w_{i}\in\mathbb{R}^{d}} \mathbb{E}_{x_i\sim\mathcal{D}_i}[\ell(f(x_i;w_{i});y_i)],
\end{equation}
for all $i=1,\dots,K$. We tackle the problem using empirical risk minimization (ERM) \cite{vapnik1999nature}, as commonly done in statistical learning setups, and in particular, we study vision classifications tasks in this paper. Meanwhile, our proposed framework is general enough to tackle any problem that fits into ERM as long as a loss function is defined.
 
\subsection{Non-iid data}
In this work, we study the problem when data is non-iid between the clients, specifically in the context of \textit{covariate-shift} and \textit{prior probability shift}. The client datasets are assumed to be private and cannot be shared, instead, the clients are allowed to communicate model parameters. In our experiments we define label shift to mean different label skews for the client data distributions, whereas covariate shift is defined as different degrees of rotation of the input images. For $K$ clients we define $N\leq K$ different data distributions $D^n(x,y)$, and divide the clients into $N$ groups defined by the $N$ distributions. Note that when $N=K$ each group is made up of a single client, resulting in all clients having different distributions of data. On the contrary, a group can also contain multiple clients and we will refer to this as a cluster. Furthermore, inter-cluster data similarity may not be uniform and some pairs of clusters may have a higher similarity than others. The client data distributions $\mathcal{D}_i^n$ are however assumed to be \textit{unknown for all clients $i$}, and each client can only see its own local data. 

Despite keeping the data private, training can be enhanced by allowing communication between the clients in the network. The training procedure, therefore, consists of each client $i$ training on the local data for $E$ local epochs and then communicating the trained model to other clients in the network. The receiving client merges its model with the received models using federated averaging and then starts a new training round of $E$ local epochs. In this work, we propose \textit{Decentralized Adaptive Clustering (DAC)} and show how this approach can find beneficial collaborations between clients in the network to better solve the local learning task.
 
\subsection{DAC: Decentralized adaptive clustering}
The intuition behind DAC is that it is beneficial for clients with similar local data distributions to communicate more frequently. For each client $i$, we use a similarity measure to compute a vector of probabilities, which is used each communication round in a weighted sampling to determine which clients $j\neq i$ to collaborate with. We use training loss as a proxy for similarity, as was previously done in \cite{ghosh2020efficient} and \cite{onoszko2021decentralized}. This similarity score is updated as the training progresses, see Figure \ref{fig:visualization} for a visualization of the algorithm. Throughout the training, we also store a history of sampled clients up until the current communication round $t$. This set is defined as $\boldsymbol{\bar{M}}_i^t = \cup_{t^{'}=0}^t\boldsymbol{M}_i^{t^{'}}$, where $\boldsymbol{M}_i^t$ is the set of sampled clients for client $i$ at communication round $t$.

For each communication round $t$, the following steps are performed for all clients $i \in 1, ... K$.
\begin{enumerate}
    \item Sample $m\leq K$ clients $\boldsymbol{M}_i^t=\{c_1,c_2,\dots,c_m\}$ without replacement with probability $\boldsymbol{p}_i^t=[p_{i1}^t, p_{i2}^t, \dots, p_{iK}^t]$, where $p_{ik}^t$ is the similarity-based probability between clients $i$ and $k$ at time $t$. At $t=0$, the sampling probability is uniform, i.e. $p^0_{ij} = \frac{1}{K}$ for all $i,j$.
    
    \item Update $p_{ik} \; \forall k \in \boldsymbol{M}_i^t$. This is done by computing a similarity score between client $i$ and $k$. The model of client $i$ is sent to client $k$, where the loss is computed training data of client $k$. We define the similarity between client $i$ and $k$ at round $t$ as the inverse of the training loss:
    \begin{equation}
    \label{eq:similarity}
        s_{ik}^t = \frac{1}{\ell(w_i^t;z_k)},
    \end{equation}
    with $\boldsymbol{s}_i = [s_{i1},s_{i2},\dots, s_{iK}]$. This vector is transformed into a probability vector for all $k$ using a softmax function
    \begin{equation}
    \label{softmax}
        \boldsymbol{p}_{i}(\boldsymbol{s}_i) = \frac{e^{\tau\boldsymbol{s}_{i}}}{\sum_{k=1}^K e^{\tau s_{ik}}}.
    \end{equation}
    
    \item In large peer-to-peer networks, it is unlikely that clients will sample all other clients in finite time. Thus, there are likely some clients that never get to calculate similarities between each other. For this reason, we define an approximated similarity score $\hat{s}^t_{ij}$ for clients $j$ that have not been sampled by $i$, but that have been sampled by clients that $i$ has sampled. In other words, we use two-hop neighbors from $i$ to estimate the approximation similarity, and it is defined as $\hat{s}^t_{ij} = s^t_{k^{*}j}$, where $k^*$ is the client in $\boldsymbol{\bar{M}}_i^{t}$ that is most similar to client $i$ and that has communicated with $j$. Formally, $k^{*} = \arg\min_{k} \boldsymbol{s}^t_{i}$ for all $k$ that have communicated with $i$ and $j$. This will make the estimation of similarities across the network more efficient as information will propagate much faster.
    
    \item Merge the model of client $i$ with the models of the sampled clients $\boldsymbol{M}_i^t$ using federated averaging.
    \item Train the merged model locally for $E$ local epochs. Repeat starting from step 1.
\end{enumerate}

\subsection{Variable DAC}
The $\tau$ parameter in equation \eqref{softmax} of DAC regulates the relative differences between the sampling probabilities. A higher $\tau$ results in a larger difference between the large and small probabilities, and in the extreme when $\tau\to\infty$ it becomes an argmax function. It is not given that a fixed $\tau$ is optimal, and we therefore also perform experiments where we let $\tau$ grow during every communication round. We start close to a random algorithm in the beginning of training, and later on, converge to more exploitation when $\tau$ increases. We name this method \textit{DAC-var}, and we increase $\tau$ every round using a sigmoid function starting at $1$ when $t=0$ and ending at a maximum value of $\tau_{\text{max}}$.

\begin{figure*}
    \centering
    \begin{subfigure}[t]{0.33\linewidth}
    \includegraphics[width=\columnwidth]{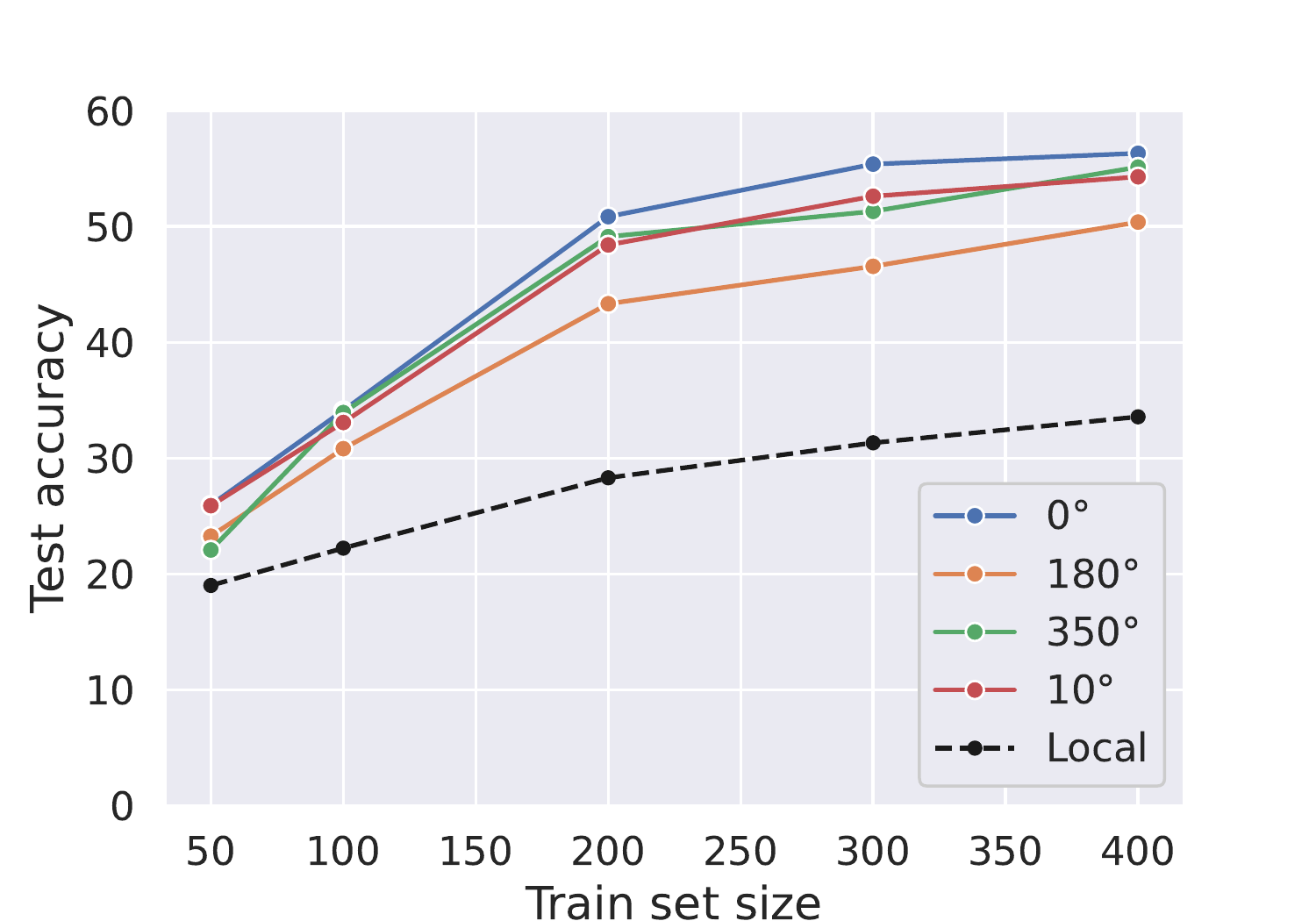}
    \subcaption{DAC}
    \label{fig:results_a}
    \end{subfigure}
    \begin{subfigure}[t]{0.33\linewidth}
    \includegraphics[width=\columnwidth]{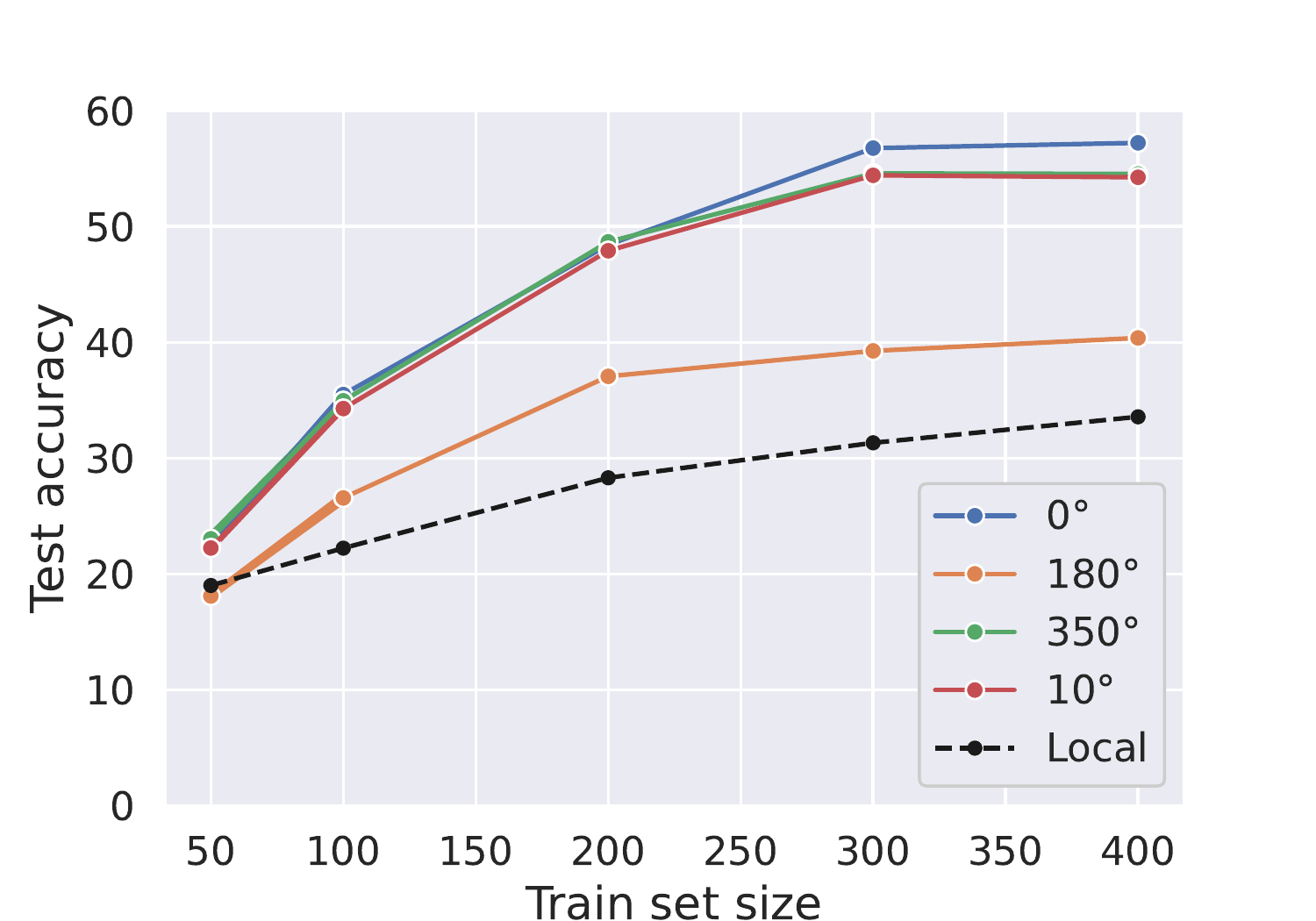}
    \subcaption{PENS}
    \label{fig:results_b}
    \end{subfigure}
    \begin{subfigure}[t]{0.33\linewidth}
    \includegraphics[width=\columnwidth]{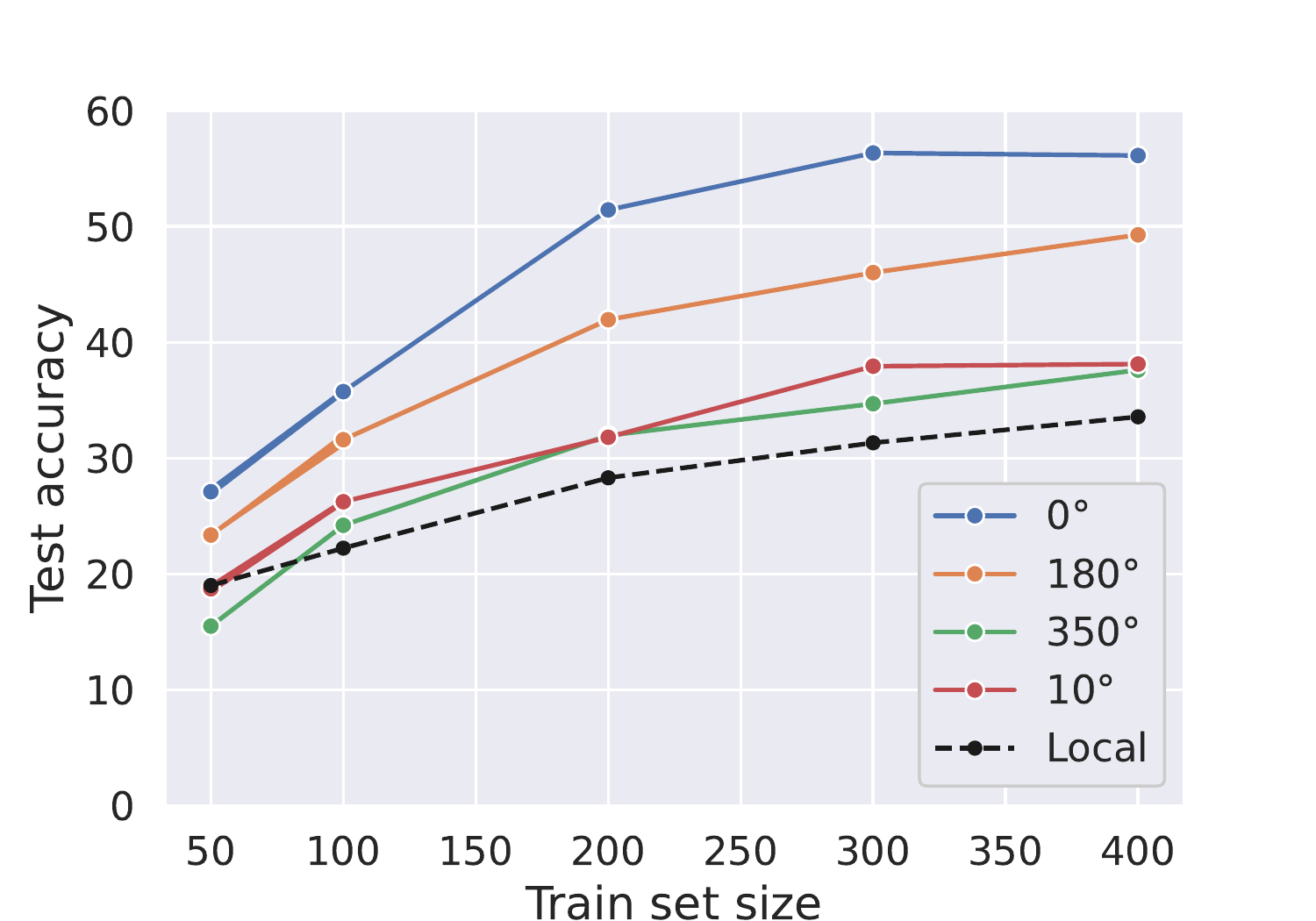}
    \subcaption{Oracle}
    \label{fig:results_c}
    \end{subfigure}
    \caption{Test accuracy on \textbf{CIFAR-10} as a function of train set size for different algorithms. The number of clients is fixed to 100, and the validation set is fixed to 100 samples.}

    \label{fig:results}
\end{figure*}

\section{Experimental setup}
We consider two main problems for decentralized learning: covariate shift and label shift. Similar to previous work, we assume that all clients can communicate with all other clients in the peer-to-peer network. For reasons of reproducibility, our code is made available on github.\footnote{\href{https://github.com/edvinli/DAC}{https://github.com/edvinli/DAC}}

\textbf{Baselines.} We compare our proposed method DAC to three baselines and an oracle. The baselines are random communication, PENS \cite{onoszko2021decentralized} and local training (without communication) on each client. The oracle has perfect cluster information and only communicates within the clusters.

\textbf{Model}. For all experiments presented in this section, we train and evaluate a 2-layer CNN with ReLU activations and max-pooling. This model is not state-of-the-art for the datasets used in our experiments, but it has sufficient capacity to solve the classification tasks in such a way that meaningful comparisons between the algorithms are possible. It is also suited for the limited amount of data available to each client in our experiments. We experiment on the CIFAR-10 \cite{krizhevsky2009learning} and Fashion-MNIST \cite{xiao2017fashion} datasets and we use cross-entropy as our loss function.

\textbf{Evaluation.} All models are evaluated on a test set drawn from the same distribution as the training set for each client in the network, that is the test set has the same (label or covariate) shift as the training set. This is since the goal is to solve the local learning task for each client as well as possible. We consider $K=100$ clients, and unless otherwise stated we distribute the number of data points uniformly over the clients. For CIFAR-10, this means 400 training samples and 100 validation samples for each client. For Fashion-MNIST it means 500 training samples and 100 validation samples. We use early stopping locally on each client, and save the model that achieved the lowest validation loss which is later used to calculate a test accuracy after $T$ communication rounds.

\textbf{Hyperparameters}. We use a fixed batch size of $8$. The Adam optimizer is used locally on each client, and learning rates are locally tuned for each dataset. For the covariate shift experiments, this results in learning rates of $\eta=3\cdot10^{-4}$ and $\eta=1\cdot10^{-5}$ for all clients on CIFAR-10 and Fashion-MNIST respectively. For the label shift experiment, a learning rate of $\eta=3\cdot 10^{-5}$ is used. The number of local epochs is fixed to $E=3$ for all experiments, and we run $T=200$ communication rounds. For PENS we use the first $20$ rounds for the neighbour selection step and then use $180$ communication rounds for cluster communication. For all algorithms each client samples and aggregates models from five clients in every communication round. For DAC, we set $\tau=30$ and for DAC-var we set $\tau_{\text{max}}=30$ for all experiments except where we analyze $\tau$. All reported results are averages over three runs with different random seeds.

We simulate a decentralized peer-to-peer network in a computer, and all models are trained using either an NVIDIA GeForce RTX 3090 Ti or an NVIDIA Tesla V100 GPU. 

\section{Results on covariate shift}
Here we summarize our findings on experiments on \textit{covariate-shift} with $K=100$ clients on both CIFAR-10 and Fashion-MNIST with equal number of data points per client. We create a setup where each client belongs to a cluster defined by rotating the input images $r$ degrees and we consider two different settings: $r\in\{0\degree,90\degree,180\degree,270\degree\}$ and $r\in\{0\degree,180\degree,10\degree,350\degree\}$. In the first case, we have four well-separated clusters and we assign equally many (25) clients to each cluster. This setting was explored in \cite{onoszko2021decentralized}. In the other setting, we create three similar clusters and one that differs more from the rest ($r=180\degree)$. Here, we assign 70, 20, 5 and 5 clients to each cluster respectively, which also makes the problem heterogeneous with respect to the cluster sizes. This setting has not been investigated in previous works.

\textbf{Local baseline.} All methods implemented in this work are also compared to local training (without any decentralized learning). For CIFAR-10, a local model trained on 400 training samples and 100 validation samples achieves on average a $33.57\%$ test accuracy, and for Fashion-MNIST a model trained on 500 training samples with 100 validation samples achieves on average a test accuracy of $73.1\%$.

The results for the first setup of rotations $\{0\degree,90\degree,180\degree,270\degree\}$ are summarized in table \ref{tab:cifar10_cov90}, and we note that our proposed method outperforms all baselines except the oracle. In table \ref{tab:cifar10_cov}, we present the results for the second setting of heterogeneous cluster sizes for both CIFAR-10 and Fashion-MNIST. The oracle is not effective in this setup, since if we only communicate inside our own (hard) cluster we will not get any benefit from other clusters which might be similar to ourselves. This is shown in our results, where the oracle only achieves around $39-40\%$ accuracy for the two small clusters. Further, we note that PENS is deteriorating in this setting, especially for the $180\degree$ cluster. This most likely is a result of the hard cluster assignment that PENS performs in its first neighbour selection step. This is further visualised in figure \ref{fig:results} which shows test accuracy as a function of training samples on each client for DAC, PENS and an oracle. 

Observing figure \ref{fig:results_c}, we see that an oracle fails to perform for the two small clusters ($10\degree$ and $350\degree$) since there are not enough clients to communicate with to learn meaningful representations, even when the number of training samples reaches 400. Here, our proposed method beats all baselines and more importantly achieves high accuracy for cluster 1 ($r=180\degree$). Further, as seen in figure \ref{fig:results}, the performance of all algorithms increases faster than that of local training since they benefit from the decentralized training, while DAC sees the largest increase in test accuracy for all clusters.

Moreover, we note in table \ref{tab:cifar10_cov} that the random baseline performs quite well in this setting, achieving a mean accuracy of $54.17\%$ over all clusters. This result stems from the fact that the largest cluster $r=0\degree$ is dominating with $70$ clients in the peer-to-peer network. Also, since cluster $r=0\degree$ is very similar to $r=350\degree$ and $r=10\degree$, the random baseline performs well also on these. However, we see that it under-performs with a test accuracy of $46.50\%$ on the $r=180\degree$ cluster, which is relatively different from the other clusters.  Meanwhile, our proposed method achieves stable accuracies over all clusters and most notably achieves a high accuracy of $49.63\%$ for the $180\degree$ cluster. Observing the results for Fashion-MNIST, the same conclusions can be drawn, but we observe an even larger difference between our proposed method and the baselines for the $r=180\degree$ cluster.

\textbf{Ablation.} An ablation experiment was performed on CIFAR-10, where we investigated the effect of the approximation similarities $\hat{s}_{ij}$ on test accuracy. This is presented as DAC - $\hat{s}_{ij}$ in table \ref{tab:cifar10_cov}, and we note that we get a higher standard deviation between the clusters without them due to lower achieved accuracy in the $180\degree$ cluster. This is most likely due to the algorithm having a hard time identifying the clusters since it only calculates the similarity between clients that communicate with each other. We thus conclude that the approximation similarity helps in cluster identification and strengthens the performance of the algorithm.

\begin{figure}[H]
    \centering
    \includegraphics[width=\columnwidth]{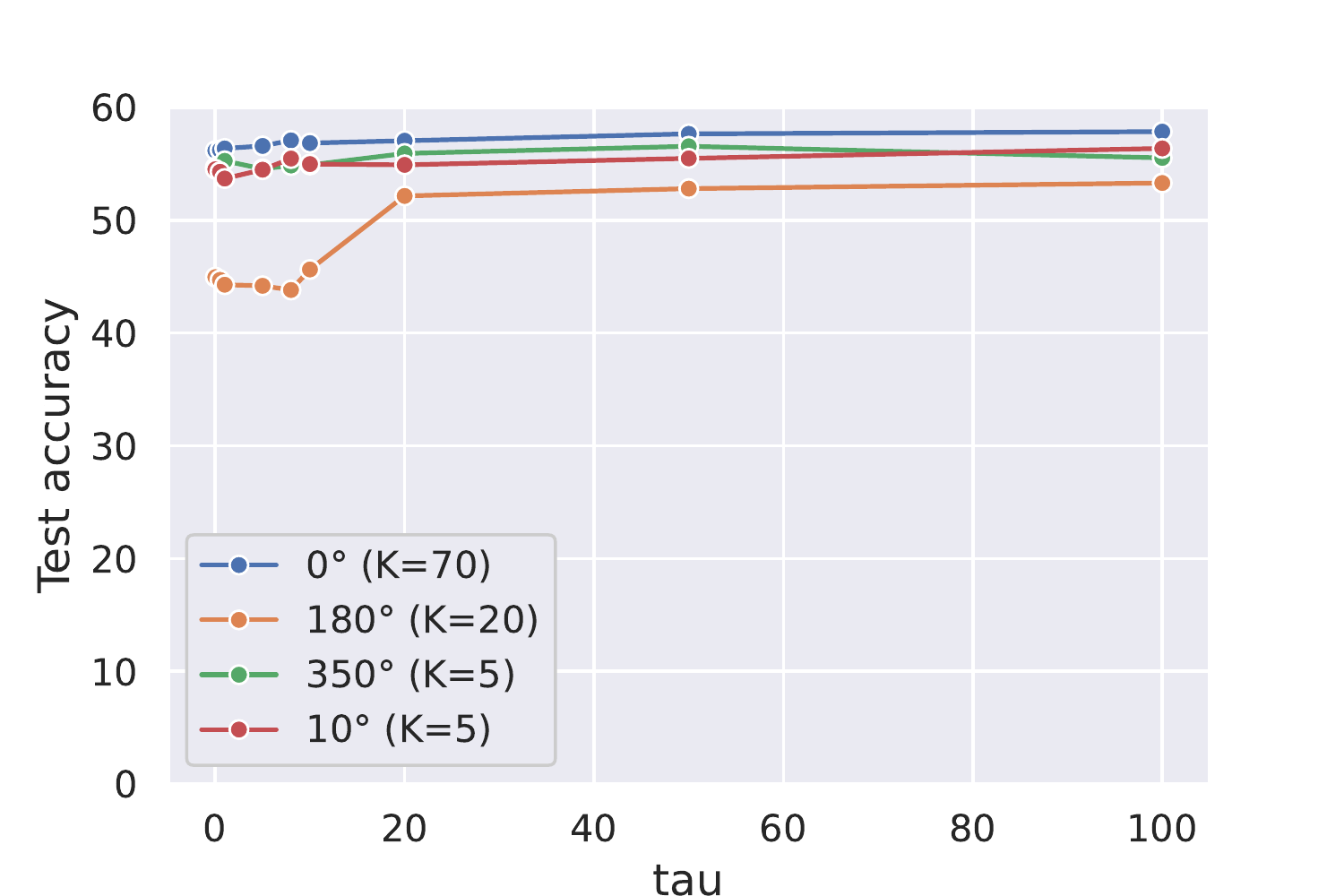}
    \caption{Test accuracy of DAC on rotated CIFAR-10 as a function of $\tau$. The random seed was fixed to the same value for all experiments.}
    \label{fig:tau}
\end{figure}

\textbf{Effect of $\tau$.} Experiments were also carried out to investigate how $\tau$ in equation \eqref{softmax} affects the performance of DAC. For this, we ran DAC on the 0-180-10-350 cluster setup while varying $\tau$. The results are summarized in figure \ref{fig:tau}. We note that for all values of $\tau$, the similar clusters are robust with respect to test accuracy. Meanwhile, for small values of $\tau$, the 'different' cluster of $r=180\degree$ achieves a relatively poor test accuracy (around $44$ to $46\%$), similar to that of the random baseline. This is expected, as low $\tau$ means that the differences between the similarity scores between clusters are small. However, as $\tau$ increases, the differences become larger and DAC is able to distinguish between clusters more efficiently which results in higher accuracy for the $180\degree$ cluster. Moreover, we note that we do not see a decrease in performance for large values of $\tau$ for any cluster.

\begin{table}[h]
\centering
\caption{Test accuracies for covariate shift on \textbf{CIFAR-10}. 25 clients in each cluster. The last column presents means over the clusters. Best mean score in bold.}
\label{tab:cifar10_cov90}
\begin{tabular}{lccccc}
\toprule
Method &   $0\degree$ &   $90\degree$ &  $180\degree$ &    $270\degree$  &    Mean \\
\midrule
DAC &  \textbf{50.27} &  \textbf{49.47} &  \textbf{51.36} &  \textbf{47.79} & \textbf{49.72}\\
Random     &  43.49 &  43.58 &  43.73 &  44.32 & 43.77 \\
PENS       &  48.30 &  49.30 &  47.69 &  47.52 & 48.20 \\ \hdashline
\textcolor{gray}{Oracle}     &  \textcolor{gray}{52.19} &  \textcolor{gray}{52.72} &  \textcolor{gray}{50.32} &  \textcolor{gray}{52.02} & \textcolor{gray}{51.81}\\ 
\bottomrule
\end{tabular}
\end{table}

\begin{table}[h]
\centering
\caption{Test accuracies for covariate shift on CIFAR-10 and Fashion-MNIST. 70, 20, 5 and 5 clients in each cluster, respectively. Last two columns present means and standard deviations over the clusters. Best mean score in bold.}
\label{tab:cifar10_cov}
\begin{tabular}{lcccccc}
\toprule
\multicolumn{7}{c}{\textbf{CIFAR-10}}\\
Method  &  $0\degree$ &        $180\degree$ &       $350\degree$ &       $10\degree$ &  Mean  & Std \\
\midrule
DAC &  $56.34$ &  $49.51$ &  $55.04$&  $54.67$ &  $53.89$ & 2.33 \\
\textbf{DAC-var} & $\textbf{57.23}$ & $\textbf{49.63}$ & $\textbf{55.57}$ & $\textbf{54.68}$ & $\textbf{54.28}$ & 2.54\\
DAC - $\hat{s}_{ij}$ & 57.12 & 46.45 & 53.77 & 55.92 & 53.31 & 4.14 \\
Random     &  $58.20$ &  $46.50$ &  $55.93$ &  $56.07$ &  $54.12$ & 4.52\\
PENS       &  57.18 &  40.56 &  $54.71$ &   $55.60$ &  $52.01$ & 5.97 \\ \hdashline
\textcolor{gray}{Oracle} &  \textcolor{gray}{$58.37$} & \textcolor{gray}{$49.99$}   &  \textcolor{gray}{$39.28$} &  \textcolor{gray}{$39.67$} &  \textcolor{gray}{$46.79$} & \textcolor{gray}{7.92} \\
\bottomrule
\multicolumn{7}{c}{\textbf{Fashion-MNIST}}\\
\textbf{DAC} &  $\textbf{75.08}$ &  $\textbf{74.95}$ &  $\textbf{73.03}$&  $\textbf{74.33}$ &  $\textbf{74.35}$ & $0.81$ \\
DAC-var & 74.48  &  74.06 & 72.58 & 73.45 &  73.64 & $0.71$ \\
Random     &  73.14 &  54.41 &  $71.02$ &  $72.52$ &  $67.77$ & $7.75$ \\
PENS       &  74.62 &  42.03 &   $73.11$ &  $73.86$ & 65.91 & $13.79$ \\ \hdashline
\textcolor{gray}{Oracle} &  \textcolor{gray}{75.78} & \textcolor{gray}{76.39}   &  \textcolor{gray}{77.99} &  \textcolor{gray}{77.77} &  \textcolor{gray}{$76.98$} & \textcolor{gray}{$0.92$}\\
\bottomrule
\end{tabular}
\end{table}

\begin{figure*}[t]
    \centering
    \begin{subfigure}[t]{0.2\linewidth}
    \includegraphics[width=1.0\columnwidth,angle=90,trim=0.5cm 5.1cm 0.5cm 2.5cm,clip=true]{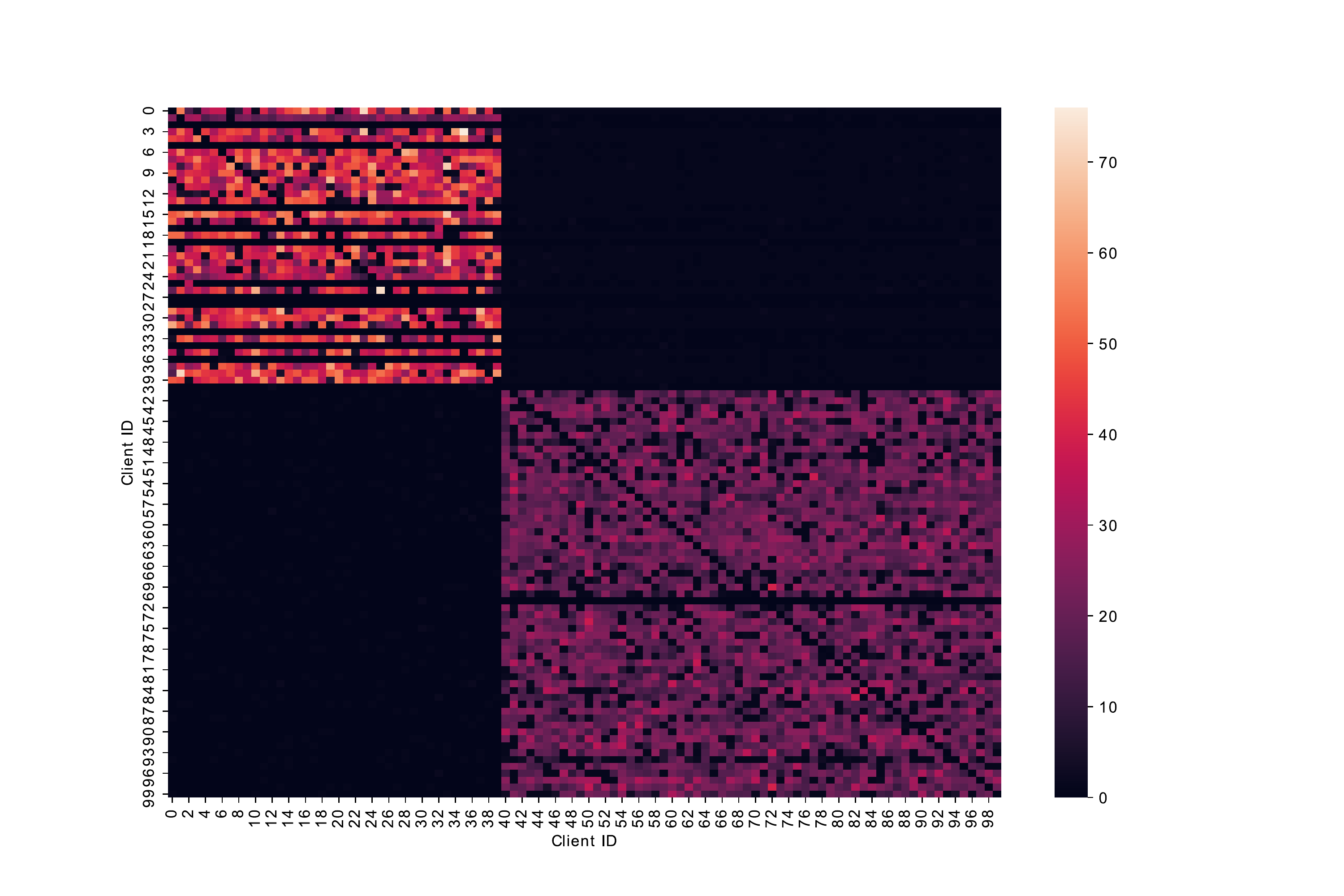}
    \subcaption{DAC}
    \label{fig:heatmaps_label_a}
    \end{subfigure}
    \hfill
    \begin{subfigure}[t]{0.2\linewidth}
    \includegraphics[width=1.0\columnwidth,angle=90,trim=0.5cm 5.1cm 0.5cm 2.5cm,clip=true]{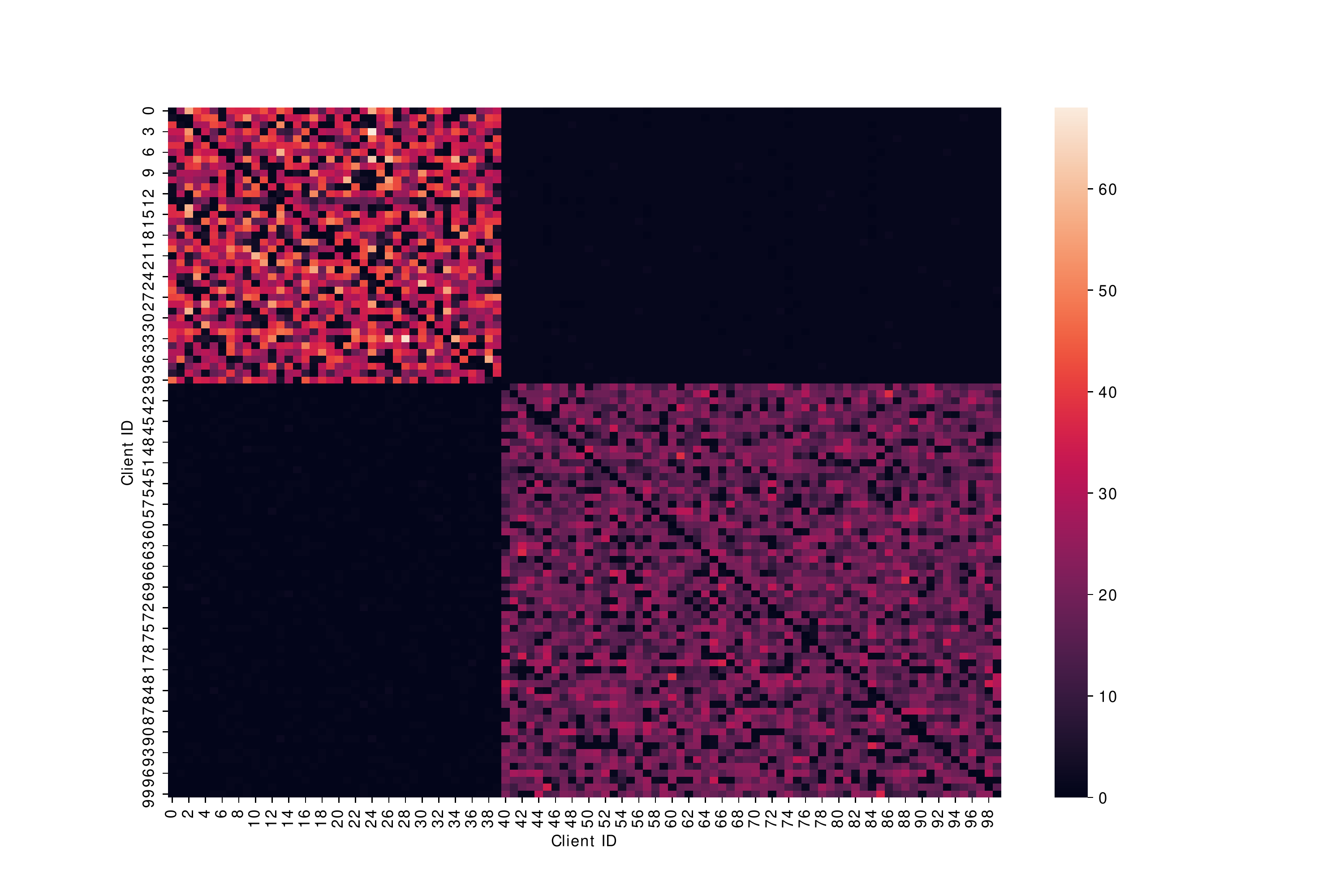}
    \subcaption{DAC-var}
    \label{fig:heatmaps_label_b}
    \end{subfigure}
    \hfill
    \begin{subfigure}[t]{0.2\linewidth}
    \includegraphics[width=1.0\columnwidth,angle=90,trim=0.5cm 5.1cm 0.5cm 2.5cm,clip=true]{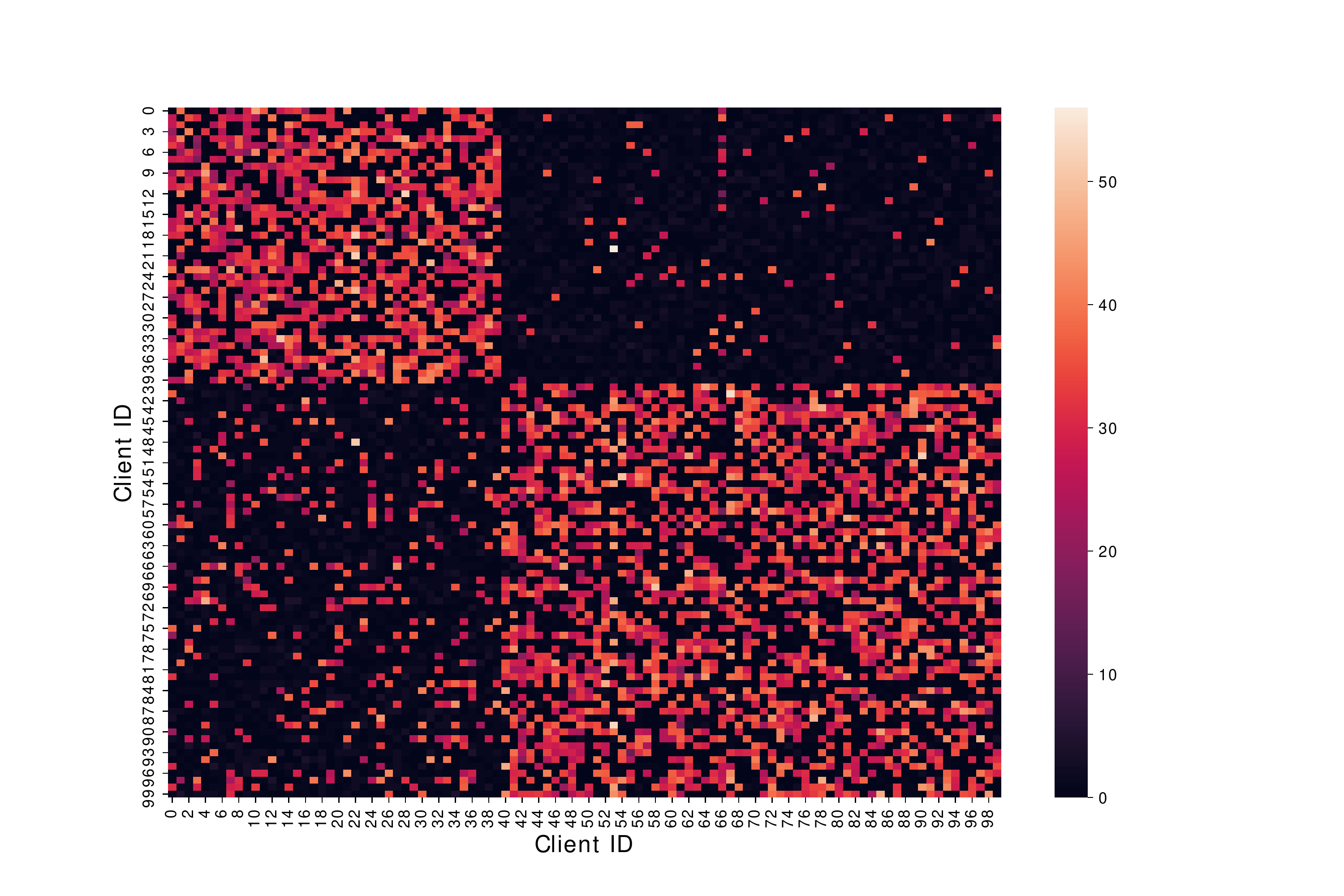}
    \subcaption{PENS}
    \label{fig:heatmaps_label_c}
    \end{subfigure}
    \hfill
    \begin{subfigure}[t]{0.2\linewidth}
    \includegraphics[width=1.0\columnwidth,angle=90,trim=0.5cm 5.1cm 0.5cm 2.5cm,clip=true]{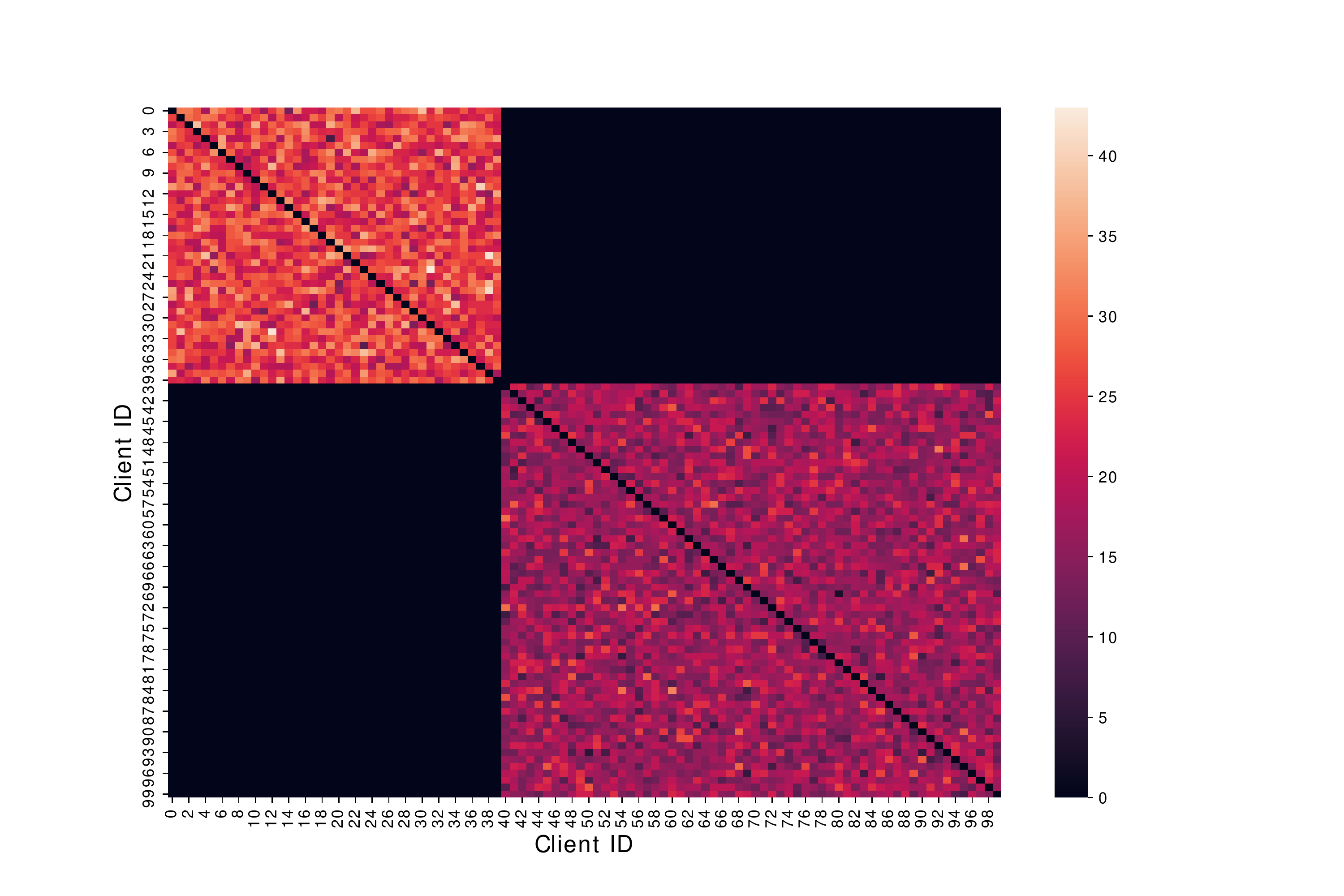}
    \subcaption{Oracle}
    \label{fig:heatmaps_label_d}
    \end{subfigure}
    \caption{Heatmaps visualising how often client $x$ communicates with client $y$ during the $T=200$ communication rounds for the label shift experiment on \textbf{CIFAR-10}. The $x$ and $y$ axes represent client IDs, which are sorted (0-39: vehicles, 40-99: animals).}
    \label{fig:heatmaps_label}
\end{figure*}
\vspace{-20pt}
\section{Results on label shift}
In this section, we present the results for experiments on label shift. This has not been studied in any of the previous works, although \cite{onoszko2021decentralized} hypothesize that PENS also should work in this setting. We study CIFAR-10, and create two different clusters using the labels: one cluster based on the animal classes (bird, cat, deer, dog, frog and horse) and one cluster based on the vehicle classes (airplane, automobile, ship and truck). We consider 100 clients, 60 in the animal cluster and 40 in the vehicle cluster. We thus have both heterogeneities in terms of cluster sizes, but also a larger distributional difference between the clusters as compared to the rotated covariate shift experiments.

The results are summarized in table \ref{tab:label}. We start by noting that the vehicles cluster is an easier task than the animals cluster since the oracle achieves a higher test accuracy on the vehicles cluster ($56.93\%$ vs $41.15\%$). We observe that both DAC and DAC-var outperform the random baseline and PENS on average, and in particular perform well on the vehicles cluster. PENS manages to perform relatively well on the animals cluster, but fails at the vehicles cluster - even performing worse than the random baseline and local training. We hypothesize that this is due to a noisy cluster assignment from the PENS neighbour identification step. Since DAC does not have any hard cluster assignments, it has the possibility to communicate with any client in the network, as opposed to PENS. This is visualised in figure \ref{fig:heatmaps_label}, which presents a heatmap over communications between clients for different algorithms (the number of times client $x$ communicates with client $y$). The $x$ and $y$ axes represent client ID:s, which are sorted with the first 40 belonging to the vehicles cluster and the last 60 to the animals cluster. We observe that DAC manages to identify clusters well, whereas PENS is much noisier.

Moreover, we see that there are horizontal artefacts for DAC in figure \ref{fig:heatmaps_label_a}. This is most likely due to the sampling getting stuck in exploiting a few clients with a high probability. Meanwhile, this is avoided for DAC-var as seen in figure \ref{fig:heatmaps_label_b}, which has a more smooth transition going from exploration in early phases to exploitation in late phases of the communication rounds. It is interesting to note that despite this, the difference in performance between DAC and DAC-var is very small.

\begin{table}[h]
\centering
\caption{Test accuracies for label shift on \textbf{CIFAR-10}. 60 clients in 'animal' cluster, and 40 clients in 'vehicle' cluster. Best mean score in bold.}
\label{tab:label}
\begin{tabular}{lcccc}
\toprule
Method  &  Vehicles & Animals &  Mean \\
\midrule
DAC & $\textbf{55.01}$ &  $\textbf{35.40}$ &  $\textbf{45.21}$ \\
DAC-var & $54.91$& $35.24$& $45.08$ \\
Random  &  $47.28$ &  $32.54$ &  $39.91$ \\
PENS  &  $30.78$ &  $35.53$ & $33.16$ \\ 
Local & 51.10 & 35.11 & 43.11 \\ \hdashline
\textcolor{gray}{Oracle} &  \textcolor{gray}{56.93} & \textcolor{gray}{41.15}  & $\textcolor{gray}{49.04}$ \\
\bottomrule
\end{tabular}
\end{table}

\section{Conclusions}
Decentralized learning is a paradigm shift in modern deep learning, as data is increasingly becoming more distributed in the world due to data collection methods and privacy concerns. We present a novel algorithm named \textit{Decentralized adaptive clustering (DAC)} for learning personalized deep nets on distributed non-iid data. Our method efficiently finds how clients in a peer-to-peer network should collaborate to solve their own local learning tasks. By sampling using a similarity metric, DAC identifies beneficial \textit{soft} clusters and leverages clients that have similar learning tasks, while at the same time keeping data private.
Our experimental results show that DAC is more efficient in finding beneficial communication patterns than previous works, as it is not limited to hard cluster assignments. Furthermore, our results show that this also leads to better performance in terms of test accuracy on image classification tasks experiencing covariate and label shift.

In this work, we have assumed that data cannot be shared due to privacy reasons. At the same time, we assume that model parameters (or equivalently, gradients) can be communicated. Previous works have shown that information about the training data can be reconstructed using gradients \cite{geiping2020inverting}. We leave this as an interesting open question to study in our framework for future work, which is compatible with methods such as differential privacy. 

Finally, we conclude that in our decentralized learning experiments label shift is a much harder problem than covariate shift. We hypothesize that this is due to aggregating models trained on widely different data distributions, which is detrimental to federated averaging. The clients thus experience so-called \textit{client drift} where the model parameters drift far away from each other in parameter space. This is further enhanced by the fact that all clients in our experiments start with differently initialized models. This aligns well with conclusions from previous works which have shown that the same initial parameters are important for federated averaging \cite{mcmahan2017communication}. Although our proposed method DAC manages to solve the problem better than both the random and local baselines, we see the problem of how to efficiently aggregate models as an interesting research question going forward.

\bibliographystyle{named}
\bibliography{biblio.bib}

\end{document}